\documentclass[10pt,twocolumn,letterpaper]{article}

\usepackage{cvpr}              

\usepackage{graphicx}
\usepackage{amsmath}
\usepackage{amssymb}
\usepackage{booktabs}

\usepackage{algorithm}
\usepackage{algorithmic}
\usepackage[dvipsnames]{xcolor}
\usepackage{multirow}
\usepackage[accsupp]{axessibility}

\usepackage[pagebackref,breaklinks,colorlinks]{hyperref}

\usepackage[capitalize]{cleveref}
\crefname{section}{Sec.}{Secs.}
\Crefname{section}{Section}{Sections}
\Crefname{table}{Table}{Tables}
\crefname{table}{Tab.}{Tabs.}

\def\cvprPaperID{3589} 
\def\confName{CVPR}
\def\confYear{2023}

\begin{document}

\title{Enhancing the Self-Universality for Transferable Targeted Attacks}

\author{Zhipeng Wei\textsuperscript{\rm 1,2}, Jingjing Chen\textsuperscript{\rm 1,2}\footnotemark[2], Zuxuan Wu\textsuperscript{\rm 1,2}, Yu-Gang Jiang\textsuperscript{\rm 1,2} \\ 
\textsuperscript{\rm 1}Shanghai Key Lab of Intell. Info. Processing, School of CS, Fudan University\\
\textsuperscript{\rm 2}Shanghai Collaborative Innovation Center of Intelligent Visual Computing
    \\
{\tt\small zpwei21@m.fudan.edu.cn}, {\tt\small \{chenjingjing, zxwu, ygj\}@fudan.edu.cn} 
}
\maketitle

\renewcommand{\thefootnote}{\fnsymbol{footnote}} 
\footnotetext[2]{Corresponding author.}

\begin{abstract}
In this paper, we propose a novel transfer-based targeted attack method that optimizes the adversarial perturbations without any extra training efforts for auxiliary networks on training data. Our new attack method is proposed based on the observation that highly universal adversarial perturbations tend to be more transferable for targeted attacks. Therefore, we propose to make the perturbation to be agnostic to different local regions within one image, which we called as self-universality. Instead of optimizing the perturbations on different images, optimizing on different regions to achieve self-universality can get rid of using extra data. Specifically, we introduce a feature similarity loss that encourages the learned perturbations to be universal by maximizing the feature similarity between adversarial perturbed global images and randomly cropped local regions. With the feature similarity loss, our method makes the features from adversarial perturbations to be more dominant than that of benign images, hence improving targeted transferability. We name the proposed attack method as Self-Universality (SU) attack. Extensive experiments demonstrate that SU can achieve high success rates for transfer-based targeted attacks. On ImageNet-compatible dataset, SU yields an improvement of 12\% compared with existing state-of-the-art methods.
Code is available at \url{https://github.com/zhipeng-wei/Self-Universality}.
\end{abstract}

\section{Introduction}
\label{sec:intro}
It has been demonstrated in recent works that adversarial examples have the properties of transferability, which means an adversarial example generated on one white-box model can be used to fool other black-box models \cite{dong2018boosting, xie2019improving, liu2020bias, wei2022towards, wei2022cross}. The existence of transferability brings convenience to performing black-box attacks, hence raising security concerns for deploying deep models in real-world applications \cite{liu2019perceptual, wei2020heuristic, zhang2021interpreting, tang2021robustart}. Consequently, considerable research attention has been spent on improving the transferability of adversarial examples for both non-targeted and targeted attacks \cite{Dong2019EvadingDT,wei2022boosting, zhao2021success}. 

Compared to non-targeted attacks, transfer-based targeted attacks are inherently much more challenging since the goal is to fool deep models into predicting the specific target class. The major difficulty of transfer-based targeted attacks is caused by the fact that the gradient directions from a source image to a target class are usually different among different DNNs \cite{liu2017delving}. Hence, transfer-based attack methods designed for non-targeted attacks typically work poorly for targeted attacks. To increase the transferability, previous studies make efforts in aligning the feature of the generated adversarial example with the feature distributions of the targeted class, which are learned from class-specific auxiliary networks \cite{inkawhich2020transferable, inkawhich2020perturbing}  or generative adversarial networks \cite{naseer2021generating}. However, these works assume that the training dataset is available and require extra training efforts for auxiliary networks, making it hard to apply in real-world scenarios. 
\begin{figure*}[t]
\centering
\includegraphics[width=0.8\textwidth]{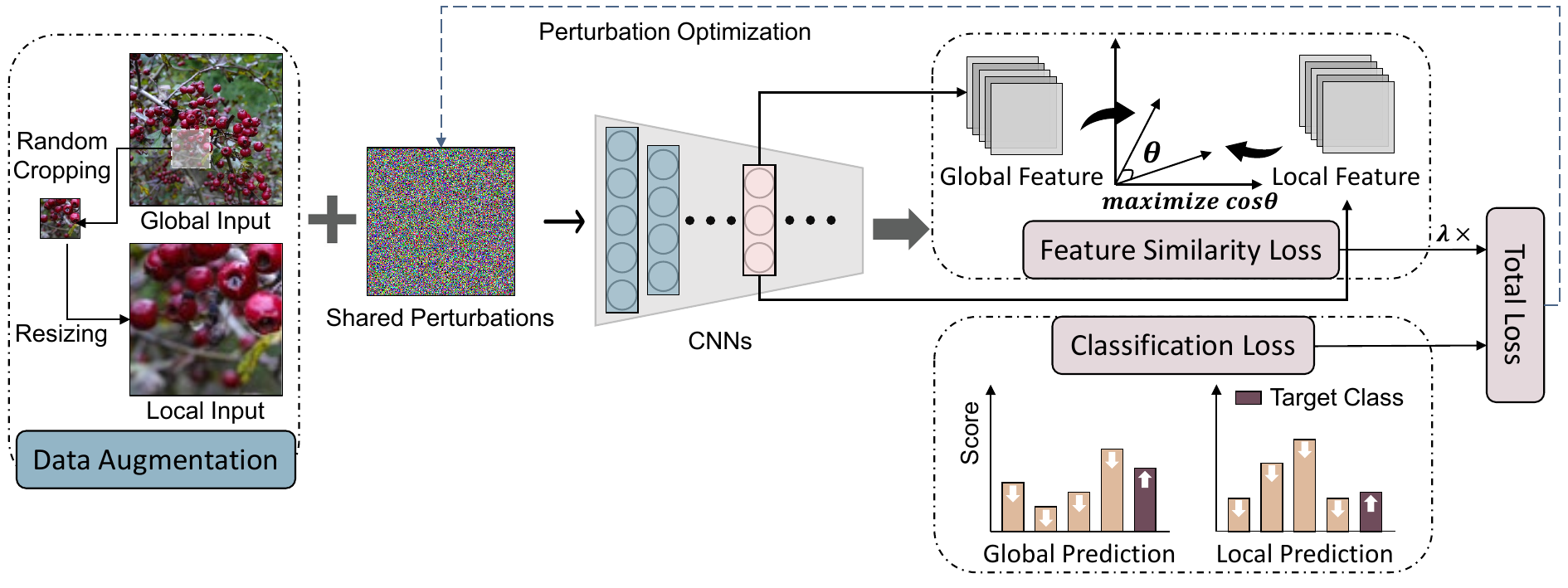}
\vspace{-1.5ex}
\caption{Overview of the proposed SU attack. The random cropping is applied to the given benign image to generate the local image patch. After cropping, the local patch is resized to the shape of the benign image. Then both benign and local adversarial images with the shared perturbations are input to a surrogate white-box CNN model. Finally, the gradients obtained from the classification loss and the feature similarity loss are used to optimize perturbations.}
\label{overview}
\end{figure*}

This paper investigates the problem of transfer-based targeted attacks. Specifically, we propose a new method that improves the transferability of adversarial examples in a more efficient way, i.e., without any training efforts for auxiliary networks to learn the feature distributions of the targeted class. Our method is proposed based on the observation that more universal perturbations yield better attack success rates in targeted attacks. To this end, our goal is to enhance the universality of the generated adversarial perturbations, in order to improve its targeted transferability. Note that existing universal adversarial perturbation (UAP) attacks \cite{MoosaviDezfooli2017UniversalAP} require optimizing the perturbations on an abundant of images to achieve universality, which is not applicable in our setting. To get rid of using extra data and make transfer-based targeted attacks as convenient as non-targeted attacks, we propose to make the perturbation to be agnostic to different local regions within one image, which we called as self-universality. Then our method optimizes the self-universality of adversarial perturbations instead. To be specific, in addition to classification loss, our Self-Universality (SU) attack method introduces a feature similarity loss that maximizes the feature similarity between adversarial perturbed global images and randomly cropped local regions to achieve self-universality. In this way, our method makes the features from adversarial perturbations to be more dominant than that of benign images, hence improving targeted transferability.

Figure \ref{overview} gives an overview of the proposed Self-Universality (SU) attack. SU firstly applies random cropping on benign images to obtain local cropped patches. Then it resizes local patches to the same size with benign images.
Consequently, global and local inputs with shared perturbations are input to the white-box model. Finally, adversarial perturbations are updated by minimizing the classification loss (\eg, Cross Entropy) between inputs and the target class and maximizing the feature similarity loss (\eg, Cosine Similarity) of adversarial intermediate features between local and global inputs.
Benefiting from satisfying the prediction of the target class between global and local inputs and approximating adversarial intermediate features between the two, the proposed SU attack can generate perturbations with self-universality, thereby improving the cross-model targeted transferability.
We briefly summarize our primary contributions as follows:
\begin{itemize}
    \item Through experiments, we find that highly universal adversarial perturbations tend to be more transferable for targeted attacks, which brings new insight into the design of transfer-based targeted attack methods.
    \item Based on the finding, we propose a novel Self-Universality (SU) attack method that enhances the universality of adversarial perturbations for better targeted transferability without the requirement for extra data. 
    \item We conduct comprehensive experiments to demonstrate that the proposed SU attack can significantly improve the cross-model targeted transferability of adversarial images. Notably, SU can be easily combined with other existing methods.
\end{itemize}

\section{Related Work}
In this section, we review existing works on non-targeted transferable attacks as well as targeted transferable attacks.
\subsection{Non-targeted Transferable Attacks}
Non-targeted transferable attacks are based on the Iterative-Fast Gradient Sign Method (I-FGSM) \cite{kurakin2018adversarial}, which iteratively calculates the gradient of the classification loss with respect to the input and updates perturbations along the direction of maximizing the loss. The multiple-step update of I-FGSM leads to more wrong predictions of white-box models. However, the generated adversarial examples are hard to attack other black-box models. This phenomenon is called over-fitting to the white-box model \cite{Kurakin2017AdversarialML}.
Subsequently, several works perform data augmentation or advanced gradient calculation to overcome the over-fitting problem. 
Data augmentation creates various input patterns, which generate adversarial perturbations with a generic pattern. For the input images, Diverse Input (DI) \cite{xie2019improving} performs random resizing and padding, Scale-invariant method (SIM) \cite{Lin2020NesterovAG} applies scale transformation, Admix \cite{wang2021admix} extends SIM by integrating images from other labels.
However, these data augmentation approaches neglect to consider more diverse input patterns (\eg, random cropping in this paper), and focus on the loss-preserving transformation \cite{Lin2020NesterovAG}. 
This is because the gradient direction fluctuates violently under more diverse input patterns without specified target classes in non-target attacks.
Advanced gradient calculation pays attention to changing gradient calculation methods or designing a new loss function.
For changing gradient calculation, the momentum term \cite{dong2018boosting} and the Nesterov accelerated gradient \cite{Lin2020NesterovAG} can be incorporated with I-FGSM to stabilize gradients.
Translation-Invariant (TI) \cite{Dong2019EvadingDT} smooths gradients through a predefined convolution kernel in order to mitigate the over-fitting problem.
Skip Gradient Method (SGM) \cite{Wu2020SkipCM} modifies the path of gradient backpropagation by skipping residual modules. Variance Tuning \cite{Wang2021EnhancingTT} utilizes gradients from data points around previous data to avoid over-fitting.
For designing a new loss function, Attention-guided Transfer Attack (ATA) \cite{Wu2020BoostingTT} and Feature Importance-aware Attack (FIA) \cite{Wang2021FeatureIT} disrupt important features that are likely to be utilized in other black-box models. ATA applies Grad-CAM \cite{Selvaraju2017GradCAMVE} to represent attention weights of features. Different to ATA, FIA computes averaged gradients of features among various augmented inputs as attention weights.
However, these new loss functions are restricted to global structures.
In contrast, this paper calculates the feature similarity loss between global and local structures of images.

\subsection{Targeted Transferable Attacks}
The transfer-based targeted attacks are more challenging than the non-targeted attacks due to the requirement of fooling the model into predicting the specified target class. Previous efforts have been devoted into training class-specific auxiliary networks \cite{inkawhich2020transferable, inkawhich2020perturbing} or generative adversarial networks (GANs) \cite{naseer2021generating}, in order to learn feature distributions of the targeted class. 
Feature Distribution Attack (FDA) \cite{inkawhich2020transferable} utilizes the intermediate features of the training dataset from the white-box model to train a binary classifier, which predicts the probability that the current feature belongs to the targeted class. Then, through learning class-wise and layer-wise feature distributions, FDA generates targeted transferable adversarial perturbations by maximizing the probability outputted from the trained binary classifier. Subsequently, $FDA^{N}+xent$ \cite{inkawhich2020perturbing} incorporates the CE loss and multi-layer information with FDA, achieving better performance. Transferable targeted perturbations (TTP) \cite{naseer2021generating} trains a generator to synthesize perturbations that seek to have similar features with targeted samples in the latent space of a pretrained discriminator.
This series of FDA works and TTP improve the performance of targeted transferable attacks, but require additional training efforts for auxiliary networks on the training dataset.

To improve the efficiency of targeted transferable attacks, recent works pay more attention to iterative attacks based on I-FGSM.
Activation Attack (AA) \cite{inkawhich2019feature} replace the classification loss with a euclidean distance loss between features of adversarial and targeted images. However, AA depends on the selected examples of the targeted class and achieves low performance on the large resolution image dataset (\eg ImageNet \cite{Deng2009ImageNetAL}). 
Different to AA, Po+Trip \cite{Li2020TowardsTT} utilizes Poincar\'{e} distance metric to solve the noise curing. They also design a triplet loss for driving adversarial examples away from the original class. However, Po+Trip evaluates the attack performance only on easy transfer scenarios, and also achieves poor performance on the targeted scenario used in the Logit attack \cite{zhao2021success}.
Logit \cite{zhao2021success} attributes the poor performance of I-FGSM based methods to the restricted number of iterations. Thus, Logit enlarges the number of iterations for ensuring the convergence of attacks. In addition, to overcome the vanishing gradient caused by the large iterations, Logit utilizes gradients of the targeted logit output with respect to inputs to update perturbations.
Different to them, our method leverages the global and local structures to generate adversarial examples with self-universality, hence improving targeted transferability. Note that the proposed SU attack can be easily combined with existing methods.

\begin{figure*}
  \centering
  \begin{subfigure}{.44\textwidth}
    \includegraphics[width=1.\linewidth]{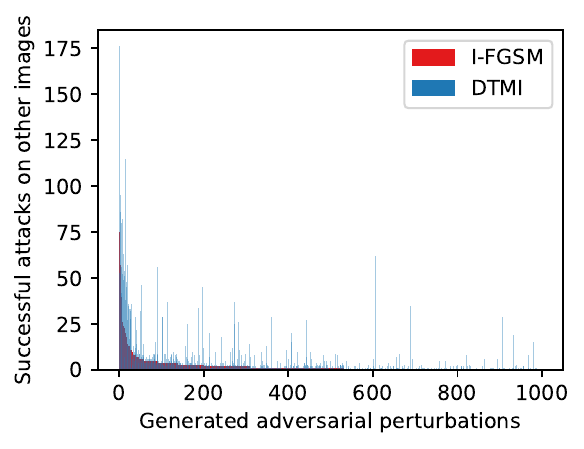}
    \caption{Universal evaluation}
    \label{fig:short-a}
  \end{subfigure}
  \begin{subfigure}{.44\textwidth}
    \includegraphics[width=1.\linewidth]{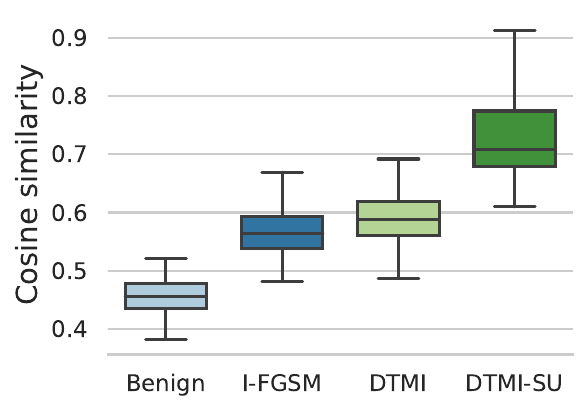}
    \caption{Feature similarity analysis}
    \label{fig:short-b}
  \end{subfigure}
  \vspace{-1.5ex}
  \caption{
(a) For each targeted perturbation generated by I-FGSM or DTMI, we calculate the number of successful targeted attacks caused by adding this perturbation to benign images other than the original optimized image.
(b) We average the cosine similarity of intermediate features among different benign images or universal adversarial images that are generated by adding the same perturbation to these benign images.} 
  \label{fig:motivation}
\end{figure*}

\section{Methodology}
\subsection{Preliminary}
Let $f$ represent the white-box surrogate model, $v$ be the black-box victim model, $x \in \mathcal{X} \subset \mathbf{R}^{H \times W \times C}$ be the benign image with the ground-truth label $y \in \mathcal{Y} = \{1, 2, ..., K\}$, where $f$ and $v$ are trained with the same training dataset, $H$, $W$, $C$ denote the number of height, width and channels respectively, and K is the number of classes. We use $f(x), v(x)$ to be the prediction over the set of classes $\mathcal{Y}$. 
Given a specified targeted class $y_{t}$, the targeted transferable attacks aim to generate adversarial examples $x_{adv}$ from the white-box model $f$ that satisfy $v(x_{adv}) = y_{t}$. Following previous works, we enforce a $L_{\infty}$-norm constraint on perturbations, which can be formulated as $||x_{adv} - x||_{\infty} = ||\delta||_{\infty} \le \epsilon$, where $\delta$ is the perturbation, $\epsilon$ denotes a constant of the norm constraint. 
Given a classification loss $J$ (\eg CE loss) of $f$, the targeted attack of I-FGSM can be formulated as:
\begin{align}
    \delta_0 &= 0,  g_{0} = 0, \label{eq:1}\\
    g_{i+1} &= \nabla_{\delta}J(f(x+\delta_i), y_t), \label{eq:2}\\
    \delta_{i+1} &= \delta_{i} - \alpha * sign (g_{i+1}), \label{eq:3}\\
    \delta_{i+1} &= Clip_{x, \epsilon}(\delta_{i+1}) \label{eq:4},
\end{align}
where $\delta_{i}$, $g_{i}$ denote the adversarial perturbation and the gradient of the $i$-th iteration, $i = [0,...,I-1]$ and $I$ is the maximum number of iterations; $\alpha$ is the step size; the function $Clip_{x,\epsilon}(\cdot)$ projects $\delta$ to the vicinity of $x$ for satisfying the $L_{\infty}$-norm constraint.
Specifically, I-FGSM first initializes $\delta_{0}$ and $g_{0}$ as $0$ (Eq.\ref{eq:1}). Then it calculates the gradients of the loss function with respect to the perturbation (Eq.\ref{eq:2}), and drives the prediction of $f$ on the adversarial example towards the targeted class by minimizing $J(f(x+\delta_i), y_t)$ (Eq.\ref{eq:3}). Finally, it restricts the adversarial perturbation with the $L_{\infty}$-norm constraint (Eq.\ref{eq:4}). Eq.\ref{eq:2}, \ref{eq:3}, \ref{eq:4} are performed alternately until the maximum iteration is reached.
Through this greedy update, I-FGSM is easily caught in local maxima and over-fitting to $f$, which results in performing poorly for targeted attacks.
Therefore, previous work \cite{zhao2021success} utilizes the combination of DI \cite{xie2019improving}, TI \cite{Dong2019EvadingDT} and MI \cite{dong2018boosting} as the baseline, since it could achieve more stable performances with a large number of iteration compared to I-FGSM. For convenience, we abbreviate the name of this combination as DTMI.
DTMI replaces Eq.\ref{eq:2} with
\begin{equation}
    g_{i+1} = \mu \cdot g_{i} + \frac{ W \cdot \nabla_{\delta}J(f(T(x+\delta_{i}, p)), y_t)}{||W \cdot \nabla_{\delta}J(f(T(x+\delta_{i}, p)), y_t)||_{1}},
\end{equation}
where $T(x+\delta_{i}, p)$ perform random resizing and padding with a probability $p$ in DI, $W$ is the pre-defined convolution kernel in TI, $\mu$ is a decay factor in MI.

\subsection{Universality of Targeted Perturbations}
Transferable targeted attacks aim to generate cross-model perturbations that drive the prediction of different models towards predicting the same specified class. It requires perturbations generated from one white-box model are model-agnostic, i.e., universal to different models. 
A simple way to meet this requirement is to optimize perturbations on massive models. However, it may be impractical in the real world, because the number of white-box models is usually limited.
In contrast, targeted universal adversarial perturbations (UAP) \cite{MoosaviDezfooli2017UniversalAP} lead to a specific prediction $y_t$ of DNNs for almost all images. \cite{Zhang2020UnderstandingAE} attributes this phenomenon to the dominant features produced by Targeted UAP. Specifically, let $\delta_{u}$ denote targeted UAP, they conclude that the predictive logit vectors of $\delta_{u}$ have a higher linear correlation with that of $x+\delta_{u}$, while the linear correlation between $x$ and $x+\delta_{u}$ is low. It suggests that the features of $\delta_{u}$ represent the feature distribution of the specific targeted class and dominate the prediction of DNNs.
Therefore, targeted UAP could transfer to attack other models trained on the same dataset, because these models share the similar feature distribution of the target class.
In summary, universality may correlate to transferability in targeted attacks.
Motivated by the above analysis, we delve into the divergence between perturbations crafted by I-FGSM and DTMI from the perspective of universality. 

To further verify the findings in the above analysis, we conduct experiments to show the correlation between universality and targeted transferability. We optimize targeted perturbations on the ImageNet-compatible dataset (dataset details are in Section \ref{exper_setting}) by I-FGSM and DTMI using DenseNet121 as the white-box model. Here CE loss is utilized as classification loss.
Given a set of benign images $\Phi=\{x^1, ..., x^m, ..., x^M\}$, we generate the perturbation for each image in the set. Denote $\delta^{m}$ as the generated perturbation for $x^m$. To evaluate the universality of $\delta^{m}$, we add $\delta^{m}$ to all benign images other than $x^{m}$, and calculate the number of successful targeted attacks on DenseNet121 by:
\begin{equation}
    \label{eq:6}
    \sum_{j=1,j \neq m}^{M} 
    \begin{cases}
    1,& \text{if } f(x^{j}+\delta^{m})=y_t,\\
    0,              & \text{otherwise.}
    \end{cases}
\end{equation}
The number of successful targeted attacks is used as the metric to evaluate the universality of $\delta^{m}$. A higher number denotes better universality. 

Figure \ref{fig:short-a} shows the comparison of universality between each perturbation crafted by I-FGSM and DTMI. For better visualization, we sort perturbations in descending order based on their universality reported in I-FGSM. 
Notably, DTMI attains higher universality than I-FGSM for each perturbation. This is because, aided by the diverse input patterns provided by DI, DTMI makes the perturbations to be universal to these input patterns.  
Since DTMI achieves better targeted transferability than I-FGSM \cite{zhao2021success}, we can basically draw a conclusion that there is a relatively positive correlation between universality and targeted transferability.

To provide insights on how the targeted perturbation affects the image features, we compare the average cosine similarity of intermediate features between benign images and adversarial images that generated by adding one specific targeted perturbation to these benign images. Specifically, these benign images can be denoted by $\Phi^{m}$, which is the set of $x^j$ satisfying $f(x^{j}+\delta^{m})=y_t$. Adversarial images can be denoted by $\{x^{j}+\delta^{m}\}$ for $x_j \in \Phi^{m}$.
Figure \ref{fig:short-b} shows the changes in the feature similarity after adding the same perturbation $\delta^{m}$. It can be observed that after adding this perturbation, the image features become more similar. This is because the targeted perturbations drive features from different benign images towards the feature distribution of the target class. In other words, compared to benign images, targeted perturbations produce more dominant features. 
In addition, from Figure \ref{fig:short-b}, we can also find that compared to I-FGSM, targeted perturbations generated by DTMI make the image features more similar. Since DTMI achieves better targeted transferability than I-FGSM, the result basically indicates that perturbations with highly targeted transferability will produce more dominant features and enjoy higher universality. 
The above analysis suggests that adversarial examples with high universality tend to be more transferable in targeted attacks.

\subsection{Self-Universality (SU) Attack}
To enhance the universality of adversarial perturbations for transferable targeted attacks, we propose the Self-Universality (SU) attack method. 
Instead of optimizing the perturbations on different images to achieve universality, our SU method optimizes the perturbation to be agnostic to different local regions within one image, which is called self-universality. In this way, our method is able to generate the universal perturbations without extra data. To achieve self-universality, our method generates perturbations by incorporating randomly cropped local regions within one image into the iterative attacks.
To create local input patterns, we consider the random cropping, which crops images into a local image patch by the scale parameter $s=\{s_l, s_{int}\}$. Where $s_l$ denotes the lower bound for the area of the random cropped images, and $s_{int}$ is the interval value between the lower and upper bounds, thus $s_l+s_{int}$ is the upper bound. After cropping images, we resize them to the same shape as benign images. Let $Loc(x, s)$ be the random cropping and the resizing operations\footnote{We perform it by RandomResizedCrop in torchvision and ignore the parameter of the random aspect ratio.}.
In addition, motivated by Figure \ref{fig:short-b}, we propose a feature similarity loss to maximize the cosine similarity of intermediate features between the adversarial global and local inputs. 
Under this design, SU could generate perturbations with more dominant features.
This loss can be formulated by $CS(f_l(x+\delta_i), f_l(Loc(x, s)+\delta_i))$, where $f_l(\cdot)$ means to extract features from the $l$-th layer of the white-box model $f$, the function $CS(\cdot, \cdot)$ calculates cosine similarity score of features between adversarial global and local inputs. 
In summary, the proposed SU replaces Eq.\ref{eq:2} with:
\begin{equation}
\begin{split}
    g_{i+1} = &\nabla_{\delta}(J(f(x+\delta_i), y_t) + J(f(Loc(x, s)+\delta_i), y_t) \\
    & - \lambda \cdot CS(f_l(x+\delta_i), f_l(Loc(x, s)+\delta_i))),
\end{split}
\end{equation}
where $\lambda$ weights the contribution of the classification loss and the proposed similarity loss.

In this way, the perturbation optimization of SU can improve the self-universality of perturbations, and the maximization of the cosine similarity can align intermediate features between adversarial global and local inputs for improving the dominance of feature representation of perturbations.
Following \cite{zhao2021success}, we integrate SU with the baseline DTMI to further boost targeted transferability, which we named DTMI-SU. We show DTMI-SU in Algorithm \ref{alg}.
In the end, SU can generate adversarial perturbations with high dominant features by incorporating local images, as shown in Figure \ref{fig:short-b}, hence improving targeted transferability.

\begin{algorithm}[tb]
\caption{DTMI-SU attack}
\label{alg}
\textbf{Input}: the classification loss function $J$, white-box model $f$, benign image $x$, targeted class $y_t$.\\
\textbf{Parameter}: The perturbation budget $\epsilon$, iteration number I, step size $\alpha$, scale parameter $s=\{s_l, s_{int}\}$, weighted parameter $\lambda$, and DTMI parameters $T(\cdot,p)$, $W$, $\mu$.\\
\textbf{Output}: The adversarial example $x_{adv}$. \\
\begin{algorithmic}[1] 
\STATE Initialize $\delta_0$ and $g_0$ by Eq.\ref{eq:1}
\FOR{$i = 0$ to $I-1$}
\STATE Random cropping and resizing: $\hat{x} = Loc(x,s)$ 
\STATE DI: $x'=T(x+\delta_i, p); \hat{x}' = T(\hat{x}+\delta_i, p)$
\STATE Calculate gradients: $g_{i+1} = \nabla_{\delta}(J(f(x'), y_t) + J(f(\hat{x}'), y_t) - \lambda \cdot CS(f_l(x'), f_l(\hat{x}')))$ \\
\STATE $g_{i+1} = \mu \cdot g_i + \frac{W\cdot g_{i+1}}{||W\cdot g_{i+1}||_1}$ \\
\STATE Update and Clip $\delta_{i+1}$ by Eq.\ref{eq:3}, \ref{eq:4}
\ENDFOR
\STATE \textbf{return} $x+\delta_{I}$ 
\end{algorithmic}
\end{algorithm}

\section{Experiment}
\subsection{Experimental Settings}
\label{exper_setting}
\subsubsection{Dataset and models}
Following \cite{zhao2021success}, we attack four diverse classifier architectures: ResNet50 \cite{He2016DeepRL}, DenseNet121 \cite{Huang2017DenselyCC}, VGGNet16 \cite{Simonyan2015VeryDC}, and Inception-v3 \cite{Szegedy2016RethinkingTI} with the ImageNet-compatible dataset\footnote{\url{https://github.com/cleverhans-lab/cleverhans/tree/master/cleverhans_v3.1.0/examples/nips17_adversarial_competition/dataset}}. 
The architectures of these four models are more diverse and challenging than the ones used in \cite{Li2020TowardsTT}. The used dataset is first introduced by the NIPS 2017 Competition on Adversarial Attacks and Defenses. It consists of 1,000 images and corresponding labels for targeted attacks.

\begin{table}[]
    \centering
    \begin{tabular}{l c c c c} \toprule
         Model & Layer 1 & Layer 2 & Layer 3 & Layer 4  \\ \midrule
         Res50 & Block-1 & Block-2 & Block-3 & Block-4\\
         Den121 & Block-1 & Block-2 & Block-3 & Block-4\\
         VGG16 & ReLU-4 & ReLU-7 & ReLU-10 & ReLU-13\\
         Inc-v3 & Conv-4a & Mix-5d & Mix-6e & Mix-7c\\ \bottomrule
    \end{tabular}
    \vspace{-1.5ex}
    \caption{Intermediate layers for feature extraction. ``Block'' means the basic block in ResNet50 and DenseNet121. ``ReLU-k'' denotes the $k$-th ReLU layer.}
    \label{tab:layer}
\end{table}

\subsubsection{Attack setting}
We use the Targeted Attack Success Rate (TASR) to evaluate the targeted attack on one black-box model, which is the rate of adversarial examples that are successfully classified as the targeted class by the black-box model. Hence, the methods with higher TASR can generate adversarial examples with high targeted transferability. Following \cite{zhao2021success}, we set the maximum perturbation as $\epsilon=16$, the step size as $\alpha=2$, the maximum number of iterations as $I=300$,
For each model, we select one layer 1/2/3/4 from shallow to deep layers (as shown in Table \ref{tab:layer}) to extract features.

\begin{table*}[t]
    \centering
    \begin{tabular}{l c c c c c c} \toprule
         \multirow{2}{*}{Attack} & \multicolumn{3}{c}{White-box Model: Res50} & \multicolumn{3}{c}{White-box Model: Dense121} \\ \cmidrule(lr){2-4}\cmidrule(lr){5-7}
         & $\rightarrow \textrm{Dense121}$ & $\rightarrow \textrm{VGG16}$ & $\rightarrow \textrm{Inc-v3}$ & $\rightarrow \textrm{Res50}$ & $\rightarrow \textrm{VGG16}$ & $\rightarrow \textrm{Inc-v3}$ \\ \midrule
         DTMI-CE & 27.1/39.7/44.3 & 18.9/27.6/29.4 & 2.2/3.4/4.1 & 12.9/16.7/18.4 & 8.1/10.6/10.6 & 1.7/2.2/3.2 \\ 
         DTMI-CE-SU & 6.2/27.8/\textbf{54.2} & 3.0/20.2/\textbf{45.4} & 0.2/4.5/\textbf{10.1} & 2.6/17.6/\textbf{39.4} & 1.4/12.6/\textbf{32.4} & 0.2/4.8/\textbf{10.8} \\ 
         \midrule
         DTMI-Logit & 30.4/64.4/71.8 & 22.6/55.1/62.8 & 2.7/7.1/9.6 & 16.1/39.3/43.7 & 13.5/33.0/38.1 & 2.1/7.1/7.7\\
         DTMI-Logit-SU & 23.8/63.9/\textbf{75.5} & 16.6/55.9/\textbf{66.9} & 2.0/8.3/\textbf{11.6} & 12.8/42.9/\textbf{50.2} & 9.3/37.2/\textbf{45.2} & 1.8/7.5/\textbf{10.4}\\ \midrule
         
         \multirow{2}{*}{Attack} & \multicolumn{3}{c}{White-box Model: VGG16} & \multicolumn{3}{c}{White-box Model: Inc-v3} \\ \cmidrule(lr){2-4}\cmidrule(lr){5-7}
         & $\rightarrow \textrm{Res50}$ & $\rightarrow \textrm{Dense121}$ & $\rightarrow \textrm{Inc-v3}$ & $\rightarrow \textrm{Res50}$ & $\rightarrow \textrm{Dense121}$ & $\rightarrow \textrm{VGG16}$ \\ \midrule
         DTMI-CE & 0.6/0.6/0.5 & 0.4/0.3/0.4 & 0.0/0.0/0.0 & 0.8/1.8/2.4 & 0.8/2.4/2.9 & 0.7/1.3/1.8 \\ 
         DTMI-CE-SU & 0.2/2.1/\textbf{2.8} & 0.2/2.1/\textbf{3.2} & 0.0/0.2/\textbf{0.2} & 0.4/1.2/\textbf{2.9} & 0.2/1.4/\textbf{5.0} & 0.1/0.8/\textbf{2.5}\\ \midrule
         DTMI-Logit & 3.0/9.6/11.3 & 3.2/12.0/13.7 & 0.1/0.6/0.7 & 0.9/2.0/2.8 & 1.1/3.3/5.0 & 0.6/2.2/3.9\\
         DTMI-Logit-SU & 3.4/11.7/\textbf{13.9} & 3.3/13.8/\textbf{16.1} & 0.2/0.9/\textbf{0.8} & 0.6/2.3/\textbf{4.3} & 0.5/3.8/\textbf{7.4} & 0.3/1.9/\textbf{4.4}\\ \bottomrule
    \end{tabular}
    \vspace{-1.5ex}
    \caption{TASR (\%) of all black-box models under four attack scenarios using ResNet50, DenseNet121, VGGNet16 and Inception-v3 as white-box models, respectively. We conduct these experiments three times and report average TASR with 20/100/300 iterations, the standard deviation is shown in Appendix. The best results with 300 iterations are in bold.
    }
    \label{tab:single-tranfer}
\end{table*}

\begin{table*}[t]
    \centering
    \begin{tabular}{l c c c c c } \toprule
         \multirow{2}{*}{Ensemble Attack} & \multicolumn{4}{c}{Black-box Model} & \multirow{2}{*}{Average}\\ \cmidrule(lr){2-5}
         & Res50 & Dense121 & VGG16 & Inc-v3 \\ \midrule
         DTMI-CE & 31.1 & 55.2 & 51.6 & 16.1 & 38.5\\
         DTMI-CE-SU & \textbf{55.7} & \textbf{65.0} & \textbf{68.2} & \textbf{29.3} & \textbf{54.5}\\ \midrule
         DTMI-Logit & 70.2 & 82.3 & 82.2 & 29.1 & 65.9\\
         DTMI-Logit-SU & \textbf{75.3} & \textbf{82.9} & \textbf{84.2} & \textbf{34.5} & \textbf{69.2} \\ \bottomrule
    \end{tabular}
    \vspace{-1.5ex}
    \caption{TASR (\%) of one black-box model in ensemble transfer attacks. TASR
    with 300 iterations is reported. The best results are in bold.}
    \label{tab:ensemble}
\end{table*}

\subsection{Performance Comparison}
Following \cite{zhao2021success}, we adopt the combination of DI, TI and MI (DTMI) as the baseline. The parameters of DTMI are the same as \cite{zhao2021success}.
We then integrate the proposed SU method with several baselines: CE loss with DTMI (DTMI-CE) and Logit loss \cite{zhao2021success} with DTMI (DTMI-Logit). We exclude Po+Trip \cite{Li2020TowardsTT} in the comparison because it has a worse performance than the Logit loss.

For SU, the weighted parameter $\lambda$ is set as $10^{-3}$, the scale parameters $s$ is set as $(0.1, 0)$, and the layer 3 is used to extract features. These parameters will be discussed in Section \ref{ablation}.

\begin{figure}[t]
\centering
\begin{subfigure}{.49\columnwidth}
    \includegraphics[width=1.\columnwidth]{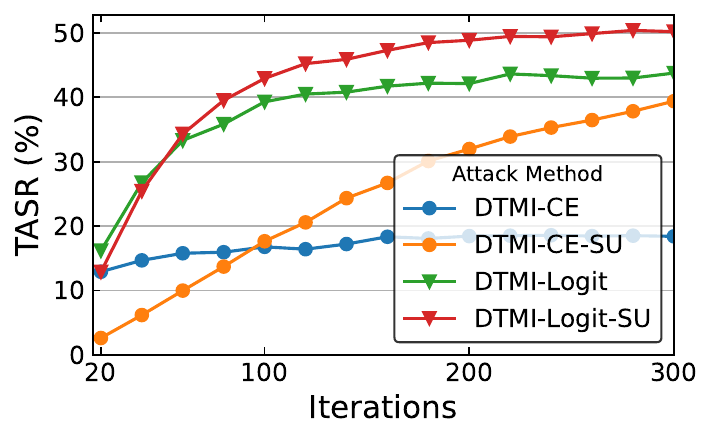}
    \caption{TASR (\%)}
    \label{fig:conver}
\end{subfigure}
\begin{subfigure}{.49\columnwidth}
    \includegraphics[width=1.\columnwidth]{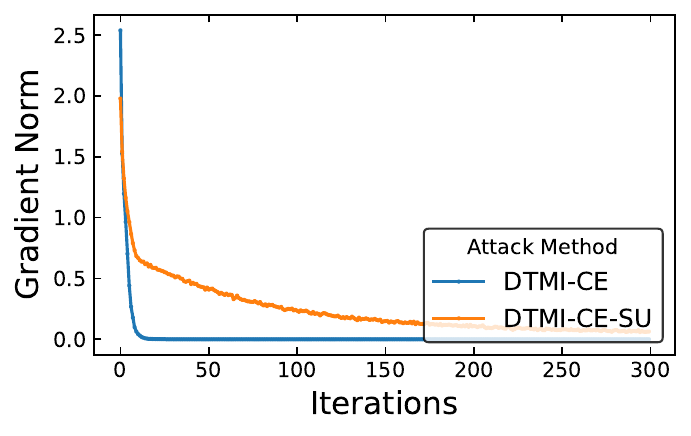}
    \caption{Gradient norm}
    \label{fig:norm}
\end{subfigure}
\vspace{-1.5ex}
\caption{(a) TASR (\%) of attacking ResNet50 from DenseNet121. (b) Gradient norm of performing DTMI-CE and DTMI-CE-SU when using DenseNet121 as the surrogate model.}
\label{fig:sin}
\end{figure}

\textbf{Single-model transferable attacks.}
We conduct single-model transferable experiments by selecting one model as the white-box model to attack the other three black-box models. 
Table \ref{tab:single-tranfer} shows the results. 
We have the following observations. 
First, the proposed SU method achieves higher TASR than DTMI-CE and DTMI-Logit by a large margin in almost all cases. It suggests that the proposed SU is suitable for different classification loss functions. 
Second, the attacks using ResNet50 and DenseNet121 as white-box models outperform that using VGGNet16 and Inception-v3, which is also observed in \cite{inkawhich2019feature, inkawhich2020transferable, zhao2021success}. It may be caused by the skip connection structure, which mitigates the gradient vanishing/exploding problem in a large number of iterations and emphasizes features of lower layers by backpropagating gradients on both residual modules and skip connections. These features tend to transfer among different models \cite{Wu2020SkipCM, wei2022cross}. 
Third, SU performs worse than other methods when using a small number of iterations (\eg $I=20$). This is because SU requires a large iteration for convergence, illustrated in Figure \ref{fig:conver}. Especially for the CE loss, DTMI achieves the convergence at $I=20$, while DTMI-SU steadily improves performance as the iteration increases, since SU can eliminate the vanishing gradient caused by CE, as shown in \ref{fig:norm}.
We also perform experiments with a smaller perturbation budget $\epsilon=8$, which are shown in Appendix. It follows a similar trend as Table \ref{tab:single-tranfer}. These results jointly demonstrate the effectiveness of SU.
In addition, we visualize adversarial examples in Appendix. These adversarial examples, which humans can correctly understand, lead to the specific target prediction of DNNs.

\textbf{Ensemble model transferable attacks.}
As mentioned above, the targeted perturbations generated from a set of models are more model-agnostic, and thus can obtain high cross-model transferability. Hence, we select one black-box model to attack, and use the other three models as white-box models. Following \cite{zhao2021success}, we simply assign equal weights to all white-box models. 
Results are shown in Table \ref{tab:ensemble}. As can be seen, compared to single-model transferable attacks, ensemble attacks achieve much better performance. It demonstrates that the model-agnostic perturbations can easily transfer to attack other models. In addition, the proposed SU further boosts cross-model targeted transferability. However, the improvement of TASR gained from DTMI-Logit-SU is not so significant. This can be explained by the fact that the loss function value of Logit increases infinitely, while the value of cosine similarity is bounded. Thus, it impedes the maximization of feature similarity. 
Nevertheless, methods that are combined with SU can promote targeted transferability in all cases.

\begin{table}[t]
    \centering
    \begin{tabular}{l c c c}
        \toprule
         Attack & Dense121 & VGG16 & Inc-v3  \\ \midrule
         DTMI-SI/+SU & 85.7/\textbf{87.2} & 69.0/\textbf{71.8} & 35.8/\textbf{41.6} \\ 
         DTMI-Adm./+SU & 89.1/\textbf{89.4} & 75.7/\textbf{79.1} & 42.1/\textbf{47.1} \\
         DTMI-EMI/+SU & 71.0/\textbf{79.0} & 64.6/\textbf{82.4} & 5.0/\textbf{14.8}\\
         ODI-TMI/+SU & 89.9/\textbf{92.8} & 81.0/\textbf{91.7} & 66.9/\textbf{72.0}\\
         \bottomrule
    \end{tabular}
    \vspace{-1.5ex}
    \caption{Average TASR (\%) of different combinational attacks. We use ResNet50 as the white-box model, and report the results with 300 iterations. Logit is used here.}
    \label{tab:la}
\end{table}

\textbf{Combination with existing methods}
To further boost transferable targeted attacks, we combine our method with other attacks, including SI \cite{Lin2020NesterovAG},  Admix \cite{wang2021admix}, EMI\cite{wang2021boosting}, and ODI \cite{byun2022improving}. Table \ref{tab:la} reports the results.
The attack methods combined with the proposed SU achieve better performance than other attack methods. In particular, compared to ODI, SU achieves an average TASR gain of 6\%. Such remarkable improvements demonstrate that the proposed SU can be easily combined with other existing methods and further improve targeted transferability.
In terms of computational efficiency, SU only requires two forward propagations for global and local inputs per iteration, which is lower than SI, Admix, and EMI (which require $\geq 5$ forward propagations). To compare SU with ODI, we report the computation time (sec) required to generate an adversarial example. DTMI-SU and DTMI-ODI require 2.3 and 2.6 sec, respectively. It indicates that SU requires less computation time than previous methods.

\begin{table}[t]
    \centering
    \begin{tabular}{c c c}
        \toprule
         Local & Feature Similarity Loss & Averaged TASR  \\ \midrule
          - & - &  9.8 \\ 
          \checkmark & - & 10.9\\ 
          \checkmark & \checkmark & 15.6\\ \bottomrule
    \end{tabular}
    \vspace{-1.5ex}
    \caption{Average TASR (\%) of black-box models for our proposed method with different component combinations. The classification loss is set as CE. `\checkmark' indicates that the component is used while `-' indicates that it is not used. TASR is averaged among four attack scenarios using ResNet50, DenseNet121, VGGNet16 and Inceptionv3 as white-box models, respectively.}
    \label{tab:aba}
\end{table}

\begin{figure}[]
\centering
\begin{subfigure}{.23\textwidth}
    \includegraphics[width=1.\columnwidth]{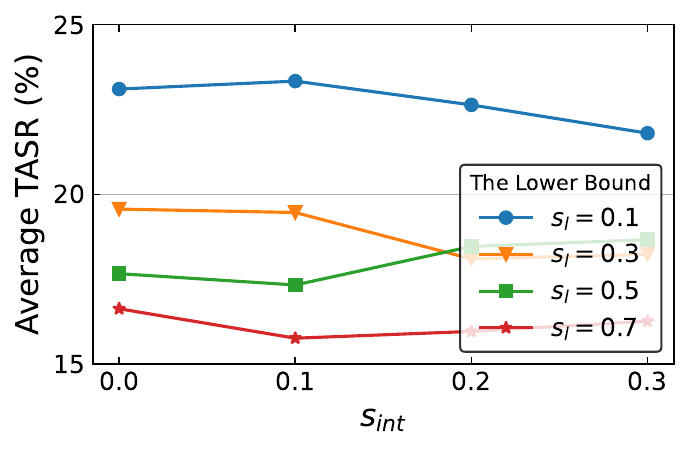}
    \caption{Scale parameter $(s_l, s_{int})$}
    \label{fig:scale}
\end{subfigure}
\begin{subfigure}{.23\textwidth}
    \includegraphics[width=1.\columnwidth]{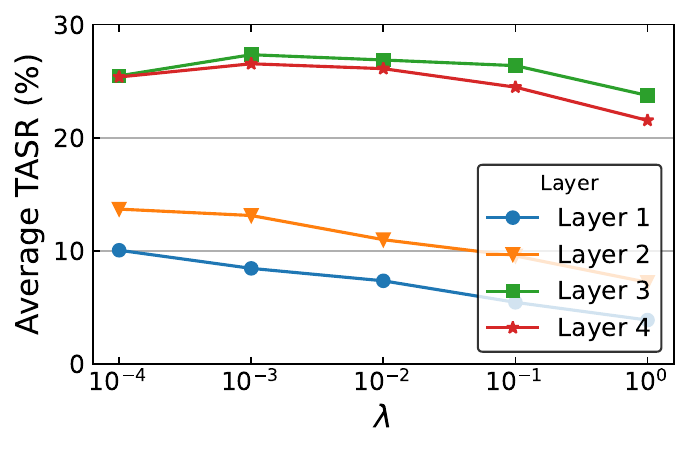}
    \caption{Layer $l$ and weight $\lambda$}
    \label{fig:layer}
\end{subfigure}
\vspace{-1.5ex}
\caption{Average TASR with different values of hyper-parameters in the attack. The classification loss is set as CE.}
\label{fig:aba}
\end{figure}

\subsection{Ablation Studies}
\label{ablation}
In this section, we utilize CE as the classification loss to investigate the effects of each component and various hyper-parameters. In addition, we investigate the effects of different regions of random cropping.

\textbf{Effect of components.}
The evaluations are conducted on four attack scenarios using ResNet50, DenseNet121, VGGNet16 and Inceptionv3 as white-box models, respectively.
We split SU into the local inputs and the feature similarity loss.
Table \ref{tab:aba} shows the results of different component combinations. As can be seen, both components improve targeted transferability. And the improvement of TASR gained from feature similarity loss is more significant.
It demonstrates the effectiveness of each component in the proposed SU attack.

\textbf{Effect of hyper-parameters.}
The evaluations are conducted by using densenet121 as the white-box model and averaging TASR among black-box models.
The scale parameter $s$ determines the area of cropped images. When the area is large, SU overlooks the local structure of inputs. Thus, it is vital to study optimal values of $s$. We fix the weighted parameter $\lambda=1.0$, and the feature extraction layer as layer 3 due to the high transferability of middle layers \cite{wei2022cross}.
Figure \ref{fig:scale} shows the results with $s_l \in [0.1, 0.3, 0.5, 0.7]$ and $s_{int} \in [0.0, 0.1, 0.2, 0.3]$. We observe that the smaller the area of cropped images, the higher the performance. It suggests that local image patches that differ significantly from the original images provide more diverse input patterns for perturbation optimization and contribute to targeted transferability.
Besides, when $s_l=0.1$ and $s_{int}=0.1$, the optimal result is reached. However, the area relative to the original image may fluctuate within $[0.1, 0.2]$. To keep it stable, we use the parameters $s=(0.1, 0.0)$ to conduct subsequent experiments.
Figure \ref{fig:layer} shows the results of performing attacks on different layers and $\lambda$. Extracting features from layer 3 is better than other layers. 
This is because the features captured by shallow layers represent the low-level patterns (such as edges), while the deepest layer contains more semantic information hence is more related to the classification task of the white-box model \cite{wei2022cross}.
For the weighted parameter $\lambda$, we set $\lambda \in [10^{-4}, 10^{-3}, 10^{-2}, 10^{-1}, 10^{0}]$. When $\lambda=10^{-3}$, SU achieves the best result.

\begin{table}[t]
    \centering
    \begin{tabular}{l c c c}
        \toprule
         Region & Res50 & VGG16 & Inc-v3  \\ \midrule
         Center & 33.0 & 29.2 & 9.9 \\ 
         Corner & 33.6 & 30.9 & 10.8 \\ 
         Whole & 39.4 & 32.4 & 10.8 \\  \midrule
         Uniform & 3.1 & 1.7 & 0.4 \\ \bottomrule
    \end{tabular}
    \vspace{-1.5ex}
    \caption{TASR (\%) of DTMI-CE-SU with different distributions of local images. The top three rows show different regions of random cropping. The last row denotes the uniform distribution $\mathcal{U}(0,1)$ of local images.}
    \label{tab:region}
\end{table}

\textbf{Effect of different regions.}
The regions of random cropping determine the distribution of local images. The center may be more related to the object of the image than the corner. Therefore, we conduct experiments on the center, the corner and the whole regions. We also compare the uniform distribution $\mathcal{U}(0,1)$ of local images. 
As shown in Table \ref{tab:region}, there is no significant performance difference between the center and the corner. The reason is that SU concentrates on the target class regardless of the object related to the original class. Therefore, directly providing more local input patterns within one image leads to better performance in SU. The whole region indeed achieves the best performance. 
Moreover, the uniform distribution makes it difficult to converge, resulting in the worst performance.

\section{Conclusion}
In this paper, we provide new insight into transfer-based targeted attacks: more universal perturbations yield better transferability.
Based on this observation, we propose the Self-University (SU) attack, which optimizes perturbations on the global image and more diverse local images, and aligns intermediate features between them. 
In this way, SU can make perturbations to be agnostic to different image regions, resulting in high self-transferability.
We conduct extensive experiments to show that SU can improve targeted transferability regardless of the single-model attack or ensemble-based attack.

\section{Acknowledgments}
This project was supported by National Key R\&D Program of China (No. 2020AAA0140001), NSFC (No. 62072116), and Science and Technology Commission of Shanghai Municipality (No. 21JC1400600).

{\small
\bibliographystyle{ieee_fullname}
\bibliography{egbib}
}

\end{document}